\documentclass{article} 
\usepackage{iclr2023_conference,times}

\usepackage{amsmath,amsfonts,bm}

\def\eqref#1{equation~\ref{#1}}

\def\1{\bm{1}}

\DeclareMathAlphabet{\mathsfit}{\encodingdefault}{\sfdefault}{m}{sl}
\SetMathAlphabet{\mathsfit}{bold}{\encodingdefault}{\sfdefault}{bx}{n}

\usepackage{hyperref}
\usepackage{url}

\usepackage{algorithm}
\usepackage{algorithmic}

\usepackage{transparent}
\usepackage{xcolor}
\usepackage{multirow}
\usepackage{graphicx}

\usepackage{graphicx}
\usepackage{caption}
\usepackage{subcaption}

\usepackage{booktabs}

\usepackage{bm}
\usepackage{mathtools}
\usepackage{bbm}

\let\norm\undefined 
\DeclarePairedDelimiter\norm{\lVert}{\rVert}

\DeclareMathOperator{\csr}{csr}

\title{SalientGrads: Sparse Models for Communication Efficient and Data Aware Distributed Federated Training}

\author{\thanks{Equal Contribution.}\, Riyasat Ohib \textsuperscript{1, 3} \quad $^{\ast}$Bishal Thapaliya \textsuperscript{2, 3} \quad Pratyush Gaggenapalli \textsuperscript{2, 3} \quad Jingyu Liu \textsuperscript{2,3}\\\textbf{Vince Calhoun \textsuperscript{1,2,3}} \quad \textbf{Sergey Plis \textsuperscript{2,3}}\\
\textsuperscript{1}Georgia Institute of Technology \quad \textsuperscript{2}Georgia State University \quad \textsuperscript{3}TReNDs Center
\\
}

\newcommand{\technique}{\textit{Salient Grads }}

\iclrfinalcopy 
\begin{document}

\maketitle
\begin{abstract}
Federated learning (FL) enables the training of a model leveraging  decentralized data in client sites while preserving privacy by not collecting data. However, one of the significant challenges of FL is limited computation and low communication bandwidth in resource limited edge client nodes. To address this, several solutions have been proposed in recent times including transmitting sparse models and learning dynamic masks iteratively, among others. However, many of these methods rely on transmitting the model weights throughout the entire training process as they are based on ad-hoc or random pruning criteria. In this work, we propose \technique which simplifies the process of sparse training by choosing a data aware subnetwork before training, based on the model-parameter's saliency scores, which is calculated from the local client data. Moreover only highly sparse gradients are transmitted between the server and client models during the training process unlike most methods that rely on sharing the entire dense model in each round. We also demonstrate the efficacy of our method in a real world federated learning application and report improvement in wall-clock communication time.
\end{abstract}

\section{Introduction}
The explosion of deep learning over the last decade has completely revolutionized entire fields including computer vision, natural language processing, recommendation systems and others. In recent times, deep learning models have continued to grow in size and with it distributed and collaborative training of such model in parallel has become a requirement. In many applications such as internet of things (IOT) and healthcare it is often the case that sensitive data is distributed in sites over great physical distance and a model needs to be trained that learns on this distributed data. It is also of paramount importance that such models are trained preserving privacy of the data without sharing it.  In many applications, data is aggregated from various organizations or devices and are pooled in a central server or a platform to train a model which is then dispersed in local sites. This becomes problematic when the data contains sensitive information. For example, to develop a neuroimaging model a series of different hospitals might want to share their data to collaboratively train the model. However, sharing patient information to a central server can reveal sensitive information and raises broad ethical concerns. A relatively recent scheme of training that tackle this setting is federated learning (FL). Federated learning is a collaborative learning technique where different devices or organizations train local models and share training information instead of sharing their data. 

Federated learning is an emerging distributed learning paradigm for decentralized data that aims to address many of the above issues including privacy \cite{mcmahan_ramage_2017}. Federated learning allows decentralized local sites to collaboratively train a shared model without sharing their local data. In the FL paradigm, a central server coordinates training process and each participating client sites (or devices) communicate only the model parameters keeping the local data private. However, in many domains and applications the data generated might be highly heterogeneous and non-IID (independent and identically distributed) \cite{zhu2021federated}. Moreover, in many scenarios the communication and the computational resources are often limited in client edge devices. Therefore, the three most pertinent challenges in the application of FL are statistical heterogeneity of the data, communication bandwidth and computational cost \cite{kairouz2021advances, li2020federated}. In this work, we aim to address the challenges of communication efficiency and computational cost in the FL setting.

\section{Related Works}
\subsection*{Federated Learning}
In the general federated learning (FL) setting, a central server tries to find a global statistical model by periodically communicating with a set $\mathcal{S}$ of clients that solve the following problem \cite{konevcny2016federated, mcmahan2017communication, bonawitz2019towards}:

\begin{align}
    \min_{x \in \mathbb{R}^d} f(x), \quad \text{where} \, f(x) = \frac{1}{N} \sum_{i=1}^{N} f_i(x)
\end{align}

Here, $N$ is the number of clients, $f_i: \mathbb{R}^d \rightarrow \mathbb{R}$ is the objective function of the local client $i$ and $f(x)$ is the global objective function. When designing a FL training paradigm a set of core considerations have to be made to maintain data privacy, address \textit{statistical} or \textit{objective} heterogeneity due to the differences in client data, and resource constraints at the client sites. A range of work tries to address the issue of heterogeneous non-IID data \cite{McMahan2016CommunicationEfficientLO, kulkarni2020survey}, however, many research also suggest that deterioration in accuracy in the FL non-IID setting is almost inevitable \cite{zhao2018federated}. In recent times, with the goal of efficient FL, effort is also being made to reduce the communication cost \cite{chen2019communication, mills2019communication, xu2020ternary}.


\subsection*{Neural Network Pruning}
Like most areas in deep learning model pruning has a rich history and mostly considered to have been explored first in the 90's \cite{janowsky1989pruning, lecun1990optimal, reed1993pruning}. The central aim of \textit{model pruning} is to find subnetworks within larger architectures by removing connections. Model pruning is very attractive due to a number of reasons, especially for real time applications on resource constraint edge devices which is often the case in FL and collaborative learning. Pruning large networks can significantly reduce the demands of inference \cite{elsen2020fast} or hardware designed to exploit sparsity \cite{cerebras, nvidia_ampere}. More recently the \textit{lottery ticket hypothesis} was proposed which predicts the existence of subnetworks of initializations within dense networks, which when trained in isolation from scratch can match in accuracy of a fully trained dense network. This rejuvenated the field of sparse deep learning \cite{renda2020comparing, chen2020lottery} and more recently the interest spilled over into sparse reinforcement learning (RL) as well \cite{arnob2021single, sokar2021dynamic}. Pruning in deep learning can broadly be classified into three categories: techniques that induces sparsity before training and at initialization \cite{lee2018snip, wang2020picking, tanaka2020pruning}, during training \cite{zhu2017prune, ma2019transformed, yang2019deephoyer, ohib2022explicit} and post training \cite{han2015deep, frankle2020pruning}. Among these for Federated Learning applications pruning at initialization holds the most promise due to selection of a subnetwork right at the start of training and the potential to only train a subset of the parameters throughout the whole training process.

\subsection*{Sparsity and Pruning in Federated Learning}

In the FL setting currently in practice, to reduce communications, it is common for the clients to perform multiple local training steps  in isolation before communicating with the server. Due to such local steps, e.g. the popular \texttt{FedAvg} algorithm suffers from a \textit{client-drift phenomenon} with objective heterogeneity, that is, the local iterates of each client drifts toward the local loss minimum and might lead to slower convergence rates \cite{li2019convergence, malinovskiy2020local, charles2020outsized, charles2021convergence}. For pruning, most of the sparse training paradigms either rely on iteratively pruning the networks that require sharing the whole model parameters each round \cite{bibikar2022federated} or sharing a sparse version of the model weights \cite{thonglek2022sparse} during training.

Relatively few research have leveraged pruning in the FL paradigm \cite{li2020lotteryfl, li2021fedmask, jiang2022model}. In particular, with LotteryFL \cite{li2020lotteryfl} and PruneFL \cite{jiang2022model}, clients need to send the full model to the server regularly resulting in higher bandwidth usage. Moreover, in \cite{li2020lotteryfl}, each client trains a personalized mask to maximize the performance only on the local data. Few recent works \cite{bibikar2022federated, huang2022achieving, qiu2022zerofl, li2020lotteryfl} also attempted to leverage sparse training within the FL setting as well. In particular \cite{li2020lotteryfl} implemented randomly initialized sparse mask, FedDST \cite{bibikar2022federated} built on the idea of RigL \cite{evci2020rigging} and mostly focussed on magnitude pruning on the server-side resulting in similar constraints. In this work, we try to alleviate these limitations which we discuss in the following section. 

\section{contributions}
In this work we propose \technique a novel paradigm for federated learning with sparse models that aims to address many of the issues with current sparse FL methods. The primary benefit of our approach is that we find sparse models to be trained using the information from the data at all the client sites and find a sparse network or a subnetwork to be trained \textit{before} the training begins. We only share the mask to all the client models \textit{once} before the training ensues and only ever transmit very sparse gradients throughout the training phase resulting in high bandwidth reduction. Since the server and all the client models start from the same initialization and have the same mask, we never need to share model weights during the training like many contemporary methods. We highlight our contributions as follows:

\begin{itemize}
    \item We propose a novel sparse Federated Learning paradigm called \technique to train only a subset of the parameters, in both server and client models, resulting in a highly communication efficient federated training technique.
    \item \technique finds a common global mask for the models based on neuron saliency scores calculated from the data in all client sites to avoid \textit{client drifts}.
    \item \technique does not need to share model parameters or masks during the training phase as we start with the same initialization and only transmits very sparse gradients to the server.
    \item We demonstrate our technique in a real world federated learning framework that trains neuroimaging models and report wall-clock time speed up.
\end{itemize}

\section{Proposed Approach}
We now introduce the \technique framework and its methodology and implementation. The principle steps in our distributed federated training mechanism are (1) Compute a model parameter saliency score based on the information from decentralized client data (2) Finding a sparse-network for the server model with the saliency score (3) training the common sub-network both in the server and the client models and communicating only the salient grads. In subsection~\ref{subsec: discovery} we elaborate the procedure to find subnetworks before training and in subsection~\ref{subsec:technique} we detail the \technique federated training method.

\begin{figure}[]
  \begin{subfigure}[b]{0.6\textwidth}
    \includegraphics[width=\textwidth]{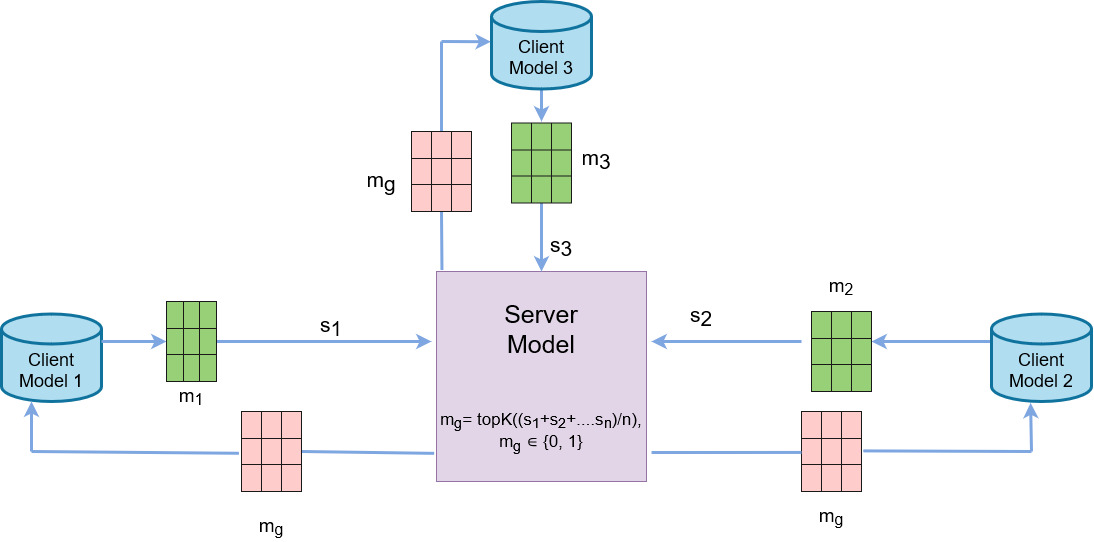}
    \caption{Before Training}
    \label{fig:f1}
  \end{subfigure}
  \hfill
  \begin{subfigure}[b]{0.3\textwidth}
    \includegraphics[width=\textwidth]{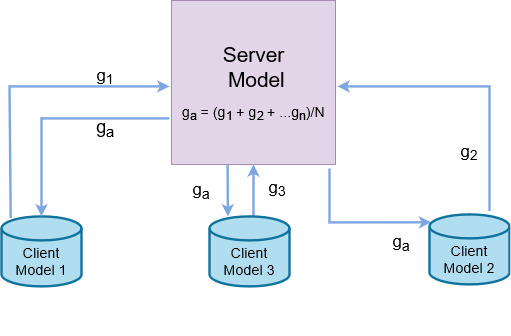}
    \caption{During Training}
    \label{fig:f2}
  \end{subfigure}
   \caption{Architecture of \technique framework(a) Before training: Server Model collects the sailency scores from client models, aggregates it and uses a \textit{topK} approach to select top 10\% of the sailency scores which creates a global mask that is shared back to all client models (b) During Training: Server Model collects the local gradients from each client model, aggregates it, and sends back the global gradients to each client models for further optimization.  }
   \label{fig:pipeline}
\end{figure}

\subsection{Discovering sub-networks at Initialization} \label{subsec: discovery}
From the \textit{Lottery Ticket Hypothesis} \cite{frankle2018lottery} we assume that there exists sub-networks of initializations within a dense network which in isolation can be trained to full accuracy of the original dense network. In the federated learning setting, the goal is to find such networks in each local sites and combining the information from the sub-networks in each of these local sites train a common model in the server. 

We consider a neural network $f$ parametrized by $\bm{\theta} \in \mathbb{R}^d$ with the parameters at initialization $\bm{\theta_0}$. The objective is to minimize the empirical risk $\mathcal{L} = \frac{1}{N} \sum_i \ell (f(\mathbf{x}_i; \bm{\theta}), y_i)$ given a training set $\mathcal{D} = \{(\mathbf{x}_i,  y_i)\}_{i=1}^N$ A sub-network within this network is defined as a sparse version of this network with a mask $\mathbf{m} \in \{0,1\}^{|\bm{\theta}|}$ that results in a masked network $f(\mathbf{x}; \bm{\theta} \odot \mathbf{m})$. 
For $N$ different client models we could initially compute $N$ different masks {$\mathbf{m}_1, \mathbf{m}_2, ... , \mathbf{m}_n$} that uniquely adapts to the local data at the client sites. However, our goal is to find a single sub-network with $\mathbf{m}_g$ leveraging the data from all the client nodes and share among all the client models once at the start of training and never again throughout the complete federated training process.
For efficiency in training and communication we want a network where $\norm{\mathbf{m}_g}_2 \ll |\bm{\theta}|$, i.e. with significantly fewer trainable parameters resulting in fewer floating point operations (FLOPs) while training and also fewer gradients to communicate between the training sites. 

The next question is how do we discover the sub-networks or masks at initialization that would result in the least amount of trade-off between accuracy and sparsity? There exists a range of neuron importance criterion in the current literature that aims to solve this problem. The most popular among these can be broadly divided into two classes: magnitude based and gradient based schemes. However, magnitude based schemes are mostly suitable for finding sub-networks post training. Hence, in our application of trying to discover sub-networks at initialization we utilize a gradient-based neuron ranking scheme. Gradient based neuron saliency or importance scheme are derived from the Taylor expansion of the variation in the loss. Some of the popular gradient based ranking criterions \cite{lee2018snip, wang2020picking, de2020progressive} can result in sparse networks at initialization that can train with reasonable loss in accuracy. Gradient based measures \cite{lee2018snip, de2020progressive} identify important connections in the network by utilizing a sensitivity measure defined as the change in the loss if the connections in question were removed. More formally, the effect of the weight $\bm{\theta}_j$ on the loss is: 
\begin{equation}
    s(\theta_j) = \lim_{ \epsilon \to 0} \left| \frac{\mathcal{L}(\bm{\theta}_0) -\mathcal{L} (\bm{\theta}_{0} + \epsilon \bm{\delta}_j)}{\epsilon} \right| = \left| \bm{\theta}_j \frac{\partial \mathcal{L}}{\partial \bm{\theta}_j} \right|
\end{equation}
where $\bm{\delta}_j$ is a vector whose $j_{th}$ element equals $\theta_k$ and all other elements are $0$. That is the saliency score for each parameter of the model is computed as the element-wise product between the parameter $\theta_j$ and its gradient $g_j$ as:
\begin{align}
    s(\theta_j) = |\theta_j \odot g_j| \label{eq:snip}
\end{align}

In the FL setting for $N$ different client sites using the saliency criterion in \ref{eq:snip}, we get the saliency score $s(\theta_0; D_k)$ for the $k^{th}$ site. To calculate the score $S_k$ we pass a few minibatches of data and average the saliency scores over the few minibatches.  To create the global mask $m_g$ we average all the saliency scores from $N$ different sites and apply the \textit{top-k} operator to find the most important connections based on the initialization and the data on all the client sites. Thus, to generate the global mask $\mathbf{m}_g \in \{0, 1\}^{|\bm{\theta}|}$ we select for the \textit{top-k} ranked connections as:
\begin{align*}
    m_g =& \mathbbm{1}[s(\theta_j) > s(\theta_k)] \\
\end{align*}
\label{eq:snip1}
where, $\theta_k$ is the $k^{th}$ largest parameter in the model and $\mathbbm{1}[\cdot]$ is the indicator function.
\subsection{Federated Learning with Salient Grads} \label{subsec:technique}
In this section we describe Federated Learning with \technique. An overview of the training flow before and during training is illustrated in the diagram Fig~\ref{fig:pipeline}. We start with a common initialization $\bm{\theta}_0$ at the server and transmit that initialization to all the client models. Next, saliency scores are calculated for each connections in the network based on the data available throughout all the clients according to the equation~\ref{eq:snip}. At this stage each different client has a different set of saliency scores for the connections in the network $f$. All the clients transmit these scores to the server where these scores are aggregated and a mask $\mathbf{m}_g$ is created corresponding to the top-k \% of the aggregated saliency scores $s_g = \sum_{i=1}^{N} s_i$. This mask is then transmitted from the server to all the client models to be used during training. The client model then trains their local models $f_i$ on the local data $\mathcal{D}_{i}$. During the training phase the client models only share their sparse masked gradients $\mathbf{g}_m = \nabla_\theta \mathcal{L}(\mathcal{D}_k) \odot \mathbf{m}_g$ to the server in the compressed sparse row (CSR) encoding where they are aggregated and transmitted to all the clients for updating the model. The algorithm for the training process is delineated in Algorithm~\ref{alg:alg-1}.

\newcommand{\loss}{\mathcal{L}}

\begin{algorithm}[]
\caption{Federate Learning with Salient Grads}
  \label{alg:alg-1}
 \begin{algorithmic}
    \STATE \texttransparent{0.5}{\# Find the Task Specific Weight Masks:}
    \STATE  $\triangleright$ Initialize model with parameters $\theta_0$.
    \STATE  $\triangleright$ transmit model to clients.
    \STATE  $\triangleright$ Verify server and all client models have the same initialization with parameters $\theta_0$.
    \STATE  $\triangleright$  $s_{1}$, $s_{2}$ .. $s_{N}$  = \text{Find neuron saliency scores} in each site using Equation \ref{eq:snip}. 
    \STATE  $\triangleright$ $\mathbf{m}_g \leftarrow T_k(S_g)$ \texttransparent{0.5}{\# calculate common mask from aggregated saliencies}
    \STATE  $\triangleright$ Server transmits $m_g$ back to all the sites.
    
\vspace{2mm}
\texttransparent{0.5}{\#In parallel do for all workers:}
  \FOR{Training steps (in parallel for all $N$ clients)}

    \STATE $\triangleright$ calculate loss $\loss(\theta \odot \mathbf{m}_g; \mathcal{D}_k)$
    \STATE $\triangleright$  $ \mathbf{g}_m \leftarrow \nabla_\theta \loss(\mathcal{D}_k) \odot \mathbf{m}_g$ \texttransparent{0.5}{\# calculate and mask gradients}
    \STATE $\triangleright$ transmit sparse gradients $\csr(\mathbf{g}_m)$ to server for aggregation
   \STATE $\triangleright$ $\mathbf{g}_a$ $\leftarrow \frac{1}{N} \sum_{m=1}^{N} g_m$ \texttransparent{0.5}{\# server aggregates all received sparse gradients $(\mathbf{g}_a)$ } 
   \vspace{1mm}
    \STATE $\triangleright$ \text{server sends aggregated gradients back to each clients} 
    \STATE $\triangleright$ $\hat{\bm{\theta}} \leftarrow \bm{\theta} - \eta \mathbf{g}_a$\texttransparent{0.5}{\# take optimization step with averaged gradients}
  \ENDFOR
\vspace{2mm}
 \end{algorithmic}
\end{algorithm}

\section{Experiments}

We performed the experiments using Amazon Web Services (AWS), by creating multiple instances to perform the federated training. Each instance has 4 cores, 16GB of memory and a Tesla M6 GPU with 6 GB of VRAM. We used a general framework called COINSTAC \cite{Plis2016}, which is a open-source federated learning solution that focuses on analysis of neuroimaging data, with an emphasis on facilitating collaboration between research institutions throughout the world. To simulate a real world FL training scenario, we selected the AWS nodes from 5 different locations throughout the world: North Virginia, Ohio, Oregan, London, and Frankfurt. We performed experiment on these five different sites, leaving additional experiments for a more exhaustive future work due to the limitation in budget and time.

\subsection{Experimental Details}
In our experiments, we consider image classification task and adopt the CIFAR-10 \cite{CIFAR10} benchmark with 60,000 samples which consists of 10 classes of 32 × 32 RGB images. For the model architecture, we train a range of ResNet models with increasing size and depth (ResNet20, ResNet32, ResNet44, ResNet56, ResNet110 and ResNet1202) with PyTorch \cite{PyTorch_NEURIPS2019_9015} to evaluate the performance of the algorithm on different scale of model parameters.  To simulate federated learning scenario, we randomly split the training set of each dataset into K batches and assign one training batch to each client. Namely, each client owns its local training, validation and testing set. At each local site, we hold out a subset of 15\% of the overall data as testing set for each local sites. Similarly, for hyperparameter tuning, we first take out a 15\% subset of training set for validation. The commonly used optimizer SGD is adopted in this experiment. Batch size of 128 was used for all the models. The learning rate was varied between  0.1 and 0.01 with a decay in learning rate by 0.1 for 50\% of total epochs and 75\% of total epochs.

The primary purpose of our experiments is to compare \technique to the standard \texttt{FedAvg} with no compression and demonstrate the viability of the method in terms of computational efficiency and performance stability. As a result, as first steps we conduct experiments using CIFAR-10 as a demonstration of the idea and initial exploration. 

\paragraph{Baseline}
We compared our model with the standard federated average model (\texttt{FedAvg}) \cite{McMahan2016CommunicationEfficientLO}. In this standard model, the gradients are calculated on each local client models, and sent to the server which aggregates and returns the gradients to all the clients for training. We will primarily focus on the performance comparison with respect to the bandwidth or the communication time, which is the time taken by the server and the clients to communicate the gradients between themselves.

\paragraph{Evaluation Metric}
The evaluation of the effectiveness of our method is mainly conducted from two perspectives: 1) communication speed and 2) model accuracy. For accuracy, we employ the average test accuracy attained on the test dataset provided by the clients. The test accuracy for local models is calculated by testing local models on the client dataset, whereas the test accuracy for global model is calculated by averaging the performance of all client models.

\subsection{Main Results}
\begin{figure}[ht]
    \centering
    \includegraphics[width= 0.8 \linewidth]{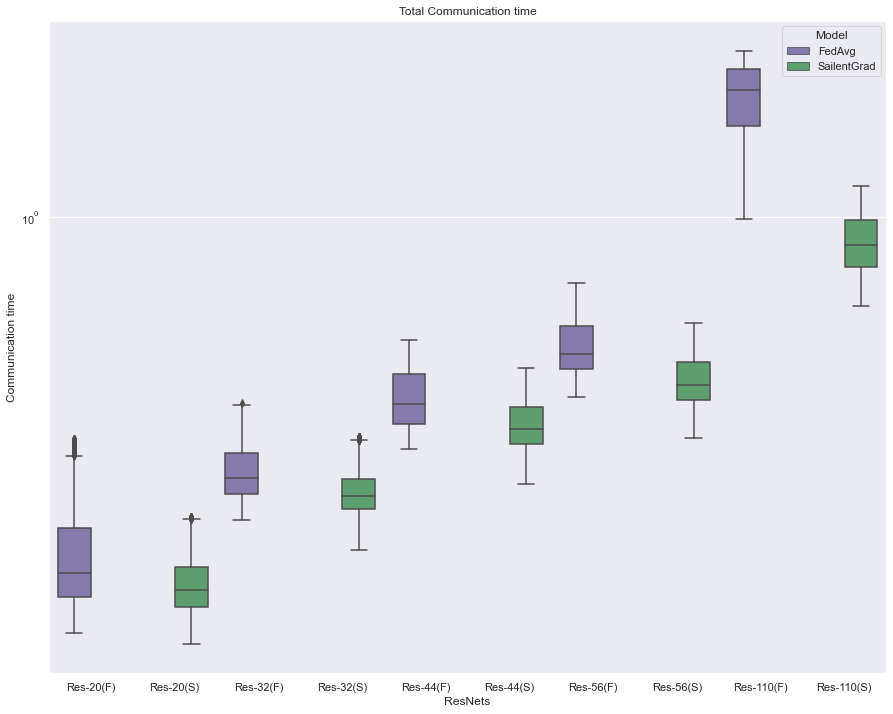}
    \caption{Performance comparision of \textit{FedAvg}VS \technique on different ResNet architectures based on communication time. }
    \label{fig:commstime}
\end{figure}

\textbf{Evaluation in standard federated learning scenarios } We first present the performance of the proposed approach on the CIFAR-10 dataset, which is evaluated in a distributed setting with 5 different local client models with varying size or depth of ResNet models. Fig~\ref{fig:commstime} shows the communication-time in seconds for \technique and \textit{FedAvg} models for different ResNet architectures in a logarithmic plot. We report the mean cumulative communication time (i.e, the time taken by the server model to gather all the gradients for each mini-batch). The sparsity for all experiments was fixed to be around 90\%, which is a relatively high sparsity in the FL setting. The average communication time between the two techniques, the total number of model parameters and the corresponding speed ups in wall-clock time are demonstrated in Table ~\ref{tab:comms_table}. Moreover, another important metric when building sparse models is the model performance. We report the model accuracy obtained by our proposed technique in Table ~\ref{tab:metrics_table} 
We also highlight the performance of our technique in terms of accuracy and similar metrics for different model architectures using CIFAR-10.

\newcommand{\lastcolnumm}{5}
\begin{table}[ht]
\centering
\makebox[1 \textwidth][c]{
\resizebox{0.80 \textwidth}{!}{
    \begin{tabular}{c|c|c|c|c}
        \toprule
        \multirow{2}{*}{\textbf{Architecture}} & \multirow{2}{*}{\textbf{Number of Parameters}} & \multicolumn{2}{c|}{\textbf{Communication Time (s)}} & \multirow{2}{*}{\textbf{Speed up}} \\ \cmidrule[0.9pt]{3-4}
        && FedAvg & Salient Grads & \\ \cmidrule[0.9pt]{1-\lastcolnumm}
        ResNet20 & 0.27M & $0.188 \pm 0.04$ & $0.147 \pm 0.04$ & $1.27$ \\\cmidrule[0.9pt]{1-\lastcolnumm}
        ResNet32 & 0.46M & $0.285 \pm 0.04$ & $0.238 \pm 0.02$ & $1.20$ \\\cmidrule[0.9pt]{1-\lastcolnumm}
        ResNet44 & 0.66M & $0.409 \pm 0.06$ & $0.328 \pm 0.04$ & $1.24$ \\\cmidrule[0.9pt]{1-\lastcolnumm}
        ResNet56 & 0.85M & $0.531 \pm 0.07$ & $0.407 \pm 0.06$ & $1.30$ \\\cmidrule[0.9pt]{1-\lastcolnumm}
        ResNet110 & 1.7M & $1.812 \pm 0.33$ & $0.781 \pm 0.13$ & $2.32$ \\
        \bottomrule
    \end{tabular}
    }
}
\caption{Performance comparison between \textit{FedAvg} and \technique on different ResNet architectures based on communication time.}
\label{tab:comms_table}
\end{table}

We observe that our \technique  framework outperforms \textit{FedAvg} for every scale of ResNet models. A clear reduction in communication time can be seen as model parameters increase, or as the models grows larger. It was observed that larger models tends to drastically benefit from our technique with almost 2.5$\times$ improvement in the communication time. In terms of accuracy, we can observe that even with 90\% sparsity, we get a stable and outstanding performance for different set of ResNet models. This is especially significant due to the real world nature of the COINSTAC framework with constrained resources and computation overheads \cite{Plis2016}.

\begin{table}[ht]
\centering
\makebox[1 \textwidth][c]{
\resizebox{0.60 \textwidth}{!}{
\begin{tabular}{c|c|c|c|c}
        \toprule
\multicolumn{1}{c}{\bf Architecture}  &\multicolumn{1}{c}{\bf Sparsity} &\multicolumn{1}{c}{\bf Accuracy} &\multicolumn{1}{c}{\bf Precision} &\multicolumn{1}{c}{\bf Recall} \\ \cmidrule[0.9pt]{1-5}
        ResNet20 & 90\% &  84.62\% & 84.48\% & 86.21\%\\\cmidrule(lr){1-5}
        ResNet32 & 90\% &  90.52\% &  90.51\% &  91.16\% \\\cmidrule(lr){1-5}
        ResNet44 & 90\% &  89.65\% &  89.61\% &  90.75\%\\\cmidrule(lr){1-5}
        ResNet56 & 90\% & 93.74\% & 93.75\% & 94.04\%\\\cmidrule(lr){1-5}
        ResNet110 & 90\% & 93.25\% & 93.20\% & 93.54\% \\
        \bottomrule
\end{tabular}
}}
\caption{Observed performance of \technique framework using CIFAR-10 dataset on different ResNet architectures}
\label{tab:metrics_table}
\end{table}

\section{Future Work}
In this work we propose a novel sparse FL method called \technique and report preliminary experimental results on the CIFAR10 dataset. We demonstrate the efficacy of our method in finding a sparse sub-network before FL training is initiated and only train a small fraction of the model parameters. We also only ever share the associated sparse gradients between the server and the clients. We leave more exhaustive analysis and detailed comparison with contemporary sparse FL methods for future works. Specifically, exploration of the possible benefits (or limitations) of this method in the non-IID setting will be explored as well as the potential to put the method in practice in a real world FL training scenario.

\section{Conclusion}
In this paper, we proposed \technique, a novel federated learning paradigm that collaboratively trains a highly sparse model without significant losses in accuracy. This framework can be effectively used to reduce the communication time and improve bandwidth during federated training, as demonstrated by the results on a range of model architectures 
There two major benefits of \technique over the existing methods 1) In most of the existing methods, all model parameters need to be shared, at least periodically, during training and as a result the communication becomes very expensive as the size of network increases. On the contrary, in \technique, only highly sparse gradients are transmitted between server and clients which significantly reduces the communication time as well as the bandwidth. 2) We compute a model parameter saliency that captures the local data characteristics at client sites and create a global model mask based on that score, resulting in a client data aware sparse model.

We also tested our method on a real world FL framework called COINSTAC, where the overall runtime of a particular algorithm can be severely constrained by the bandwidth of distributed learning algorithms. Our initial investigation on our novel sparse FL technique has revealed improvements on communication time which makes the framework excel in bandwidth limited settings without any significant accuracy loss. In future works, we aim to conduct more elaborate experiments with more computational and experimental constraints akin to real-world scenario.



\subsubsection*{Acknowledgments}
This work was supported by NIH R01DA040487 and in part by NSF 2112455, and NIH 2R01EB006841

\bibliography{ref}
\bibliographystyle{iclr2023_conference}


\end{document}